\documentclass[11pt]{article}
\usepackage{amsmath,amsthm,amsfonts,amssymb}
\usepackage[margin=1in]{geometry}
\usepackage{fancyhdr}
\usepackage{hyperref}
\usepackage{xcolor}
\usepackage{tcolorbox}
\usepackage{graphicx}
\usepackage{subcaption}
\usepackage{tcolorbox}
\usepackage{enumerate}
\usepackage{tikz}
\usepackage{hyperref}
\usepackage{enumitem}

\usepackage{algorithmic}
\title{\vspace{-2.0cm}\textbf{Multi-step Inference over Unstructured Data}}
\author{
Aditya Kalyanpur\footnote{Corresponding Author: adityak@ec.ai}, Kailash Karthik Saravanakumar, Victor Barres, \\
CJ McFate, Lori Moon, Nati Seifu, Maksim Eremeev, \\
Jose Barrera, Abraham Bautista-Castillo, Eric Brown, David Ferrucci\\
\\
Elemental Cognition Inc.\\
}

\date{}

\begin{document}

\maketitle

\abstract{The advent of Large Language Models (LLMs) and Generative AI has revolutionized natural language applications across various domains. However, high-stakes decision-making tasks in fields such as medical, legal and finance require a level of precision, comprehensiveness, and logical consistency that pure LLM or Retrieval-Augmented-Generation (RAG) approaches often fail to deliver. At Elemental Cognition (EC), we have developed a neuro-symbolic AI platform to tackle these problems. The platform integrates fine-tuned LLMs for knowledge extraction and alignment with a robust symbolic reasoning engine for logical inference, planning and interactive constraint solving. We describe Cora, a Collaborative Research Assistant built on this platform, that is designed to perform complex research and discovery tasks in high-stakes domains. This paper discusses the multi-step inference challenges inherent in such domains, critiques the limitations of existing LLM-based methods, and demonstrates how Cora’s neuro-symbolic approach effectively addresses these issues. We provide an overview of the system architecture, key algorithms for knowledge extraction and formal reasoning, and present initial evaluation results that highlight Cora’s superior performance compared to well-known LLM and RAG baselines.}

\section{Introduction}

With the emergence of Large Language Models (LLMs) and Generative AI, there is an enormous interest in building natural language applications for a wide variety of use-cases across multiple domains. Gen-AI is being leveraged in solutions ranging from conversational web search and enterprise search engines, to chat-bots for customer service, retail, travel, insurance, etc.

There is a class of high stakes decision-making applications that require performing accurate, detailed and well rationalized research to evaluate and justify complex hypotheses. Such applications include Life Science and Medical research for drug discovery and Macro-economic analysis for investment research. These use-cases are challenging to tackle using pure LLM or even Retrieval-Augmented-Generation (RAG, which is LLM+search) based approaches, due to the need to be precise, thorough and logically consistent. 

In particular, the use-cases have the following characteristics:
\begin{itemize}
    \item There are highly adverse effects to being wrong -- in some cases, lives are at stake, in others, there is a risk in losing enormous time, money and resources to pursuing a wrong path. The explicability and transparency of the system are therefore critical, as decision makers need strong and precise evidence to justify their actions. 
    \item The problems involve exploring and evaluating complex research hypotheses where the answers and evidence are often not specified in a single source or document; instead, relevant information is spread across multiple datasets, which need to be pieced together.
    \item The underlying data is a mix of large unstructured text corpora (e.g., PubMed\footnote{https://pubmed.ncbi.nlm.nih.gov/}, Financial news articles) and structured data, ontologies and knowledge graphs (e.g., Gene Ontology \cite{gene-ontology}). Combined with the above point, it means we need to extract knowledge (key concepts and relationships) from unstructured data, link it to relevant structured data, and build a \emph{unified} knowledge source to help connect the dots when exploring hypotheses.
    \item Finding \emph{refuting} evidence is as crucial as supporting evidence in order to provide a fully balanced view of the problem, and avoid confirmation bias.
    \item Developing a strong understanding of the problem space and building sufficient confidence in the solution requires causal and logical inference over multiple inter-dependent causal factors and linkages. We need to develop reasoning strategies that help users probe and clarify assumptions, detect and explain contradictions in the data, do probabilistic analysis weighing supporting and refuting evidence appropriately, and support counterfactual reasoning (\emph{What-if} analysis).
\end{itemize}

At Elemental Cognition (EC), we have built solutions to tackle these problems that combine fine-tuned LLMs (and in some cases, smaller transformer based pre-trained LMs that are more optimal for a given task) for knowledge extraction, alignment and fluent NL generation, with our multi-strategy Symbolic Reasoning Engine for precise logical reasoning and constraint solving. The technologies we have built (components, models, APIs) are part of our Neuro-Symbolic AI platform (described in Section \ref{sec:platform}). To showcase the platform's potential, we have developed an application called \emph{Cora} (Collaborative Research Assistant) that uses various platform APIs for knowledge retrieval, synthesis and reasoning, and is designed to perform complex research and discovery in high-stakes problem domains. 

In this paper, we start by describing the multi-step inference problems and discuss the challenges a pure statistical/LLM-based approach faces when solving them. We then show how EC's Cora resolves these challenges using a neuro-symbolic approach. We provide an overview of the underlying technology and system architecture, and highlight key algorithms for knowledge extraction and formal reasoning. Finally, we evaluate our system against well-known LLM and RAG baselines on the multi-step inference problems to demonstrate their value.

\section{Multi-Step Inference Use-Cases}

\subsection{Life Science Research: Drug Discovery and Re-purposing}

\begin{figure}[htb]
\centering
\includegraphics[width=0.5\columnwidth]{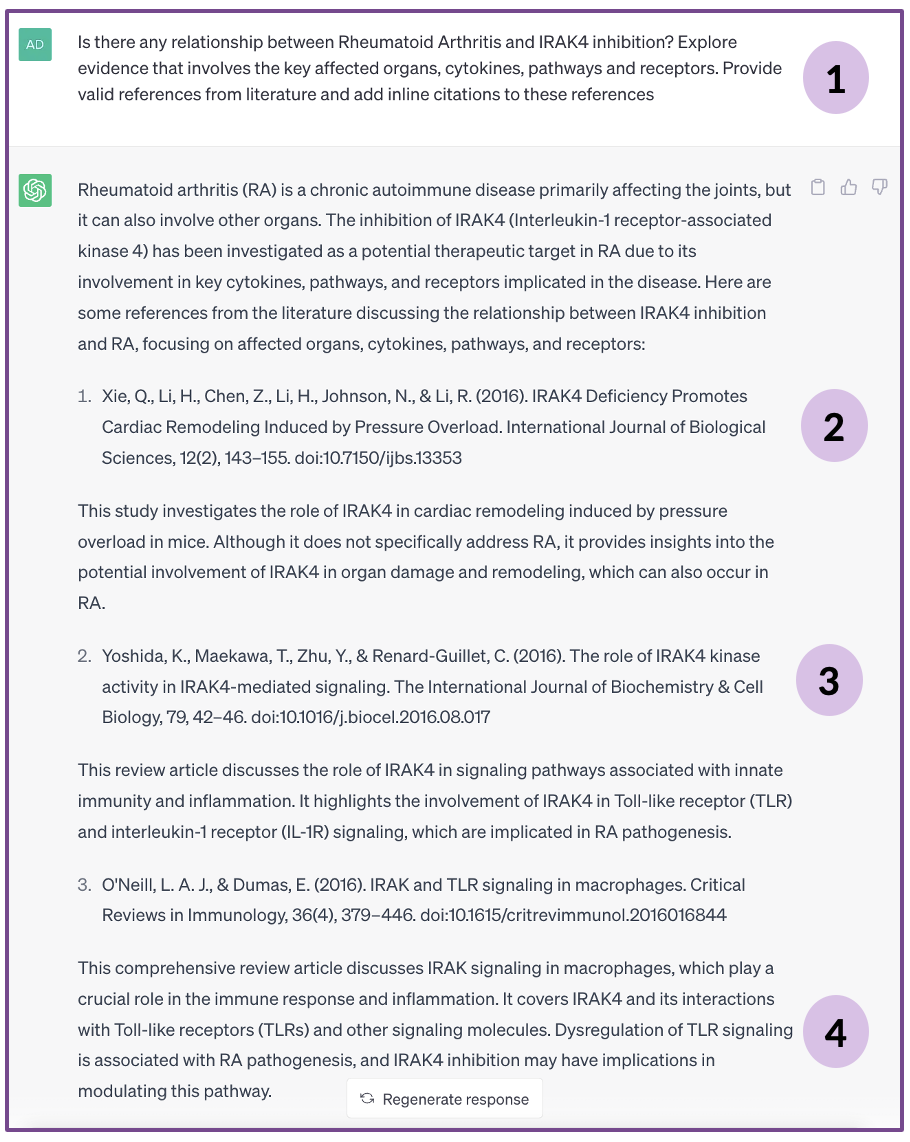}
\caption{\textbf{Using ChatGPT for Medical Research.} There are four main classes of problems (1) No control over the search process, filtering or ranking of results; (2) Inability to validate without cross-checking references - here, the paper exists but it does not contain evidence justifying the claim; (3) Hallucinated references - this citation is made up; (4) Cannot guarantee completeness - inability to find needles in the haystack}
\label{fig:gpt-medical}
\end{figure}

\begin{figure}[htb]
\centering
\includegraphics[width=0.5\columnwidth]{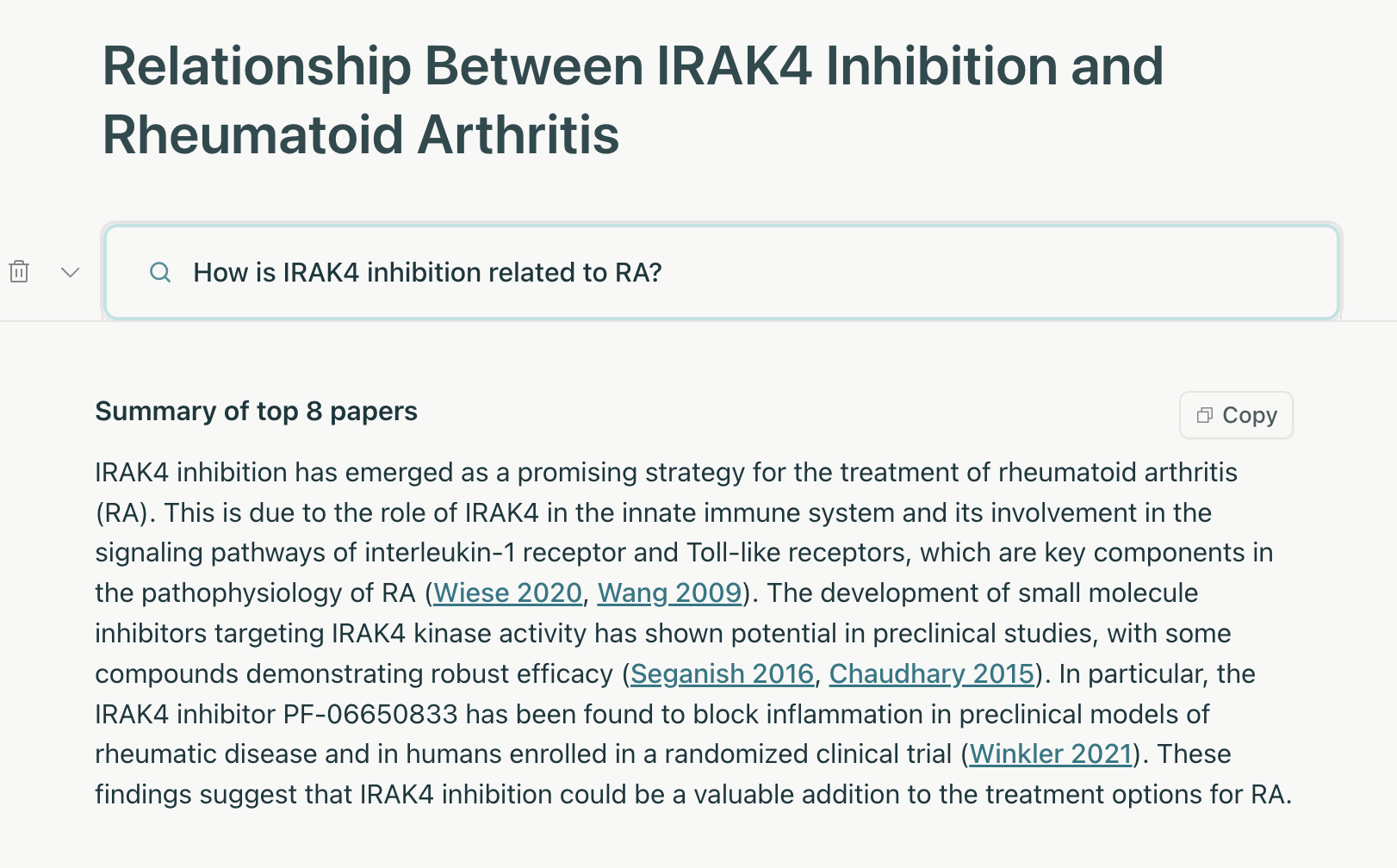}
\caption{\textbf{Elicit's answer to the question linking IRAK4 and RA}}
\label{fig:elicit-qa}
\end{figure}

A biopharma company can spend over \$2 billion to take a drug from initial discovery to approved use in the market \cite{wouters2020costs}. Identifying high-quality drug targets at the beginning of this process is essential both to increase the chance of success and to reduce the downstream costs. Comprehensive and fast literature review at the earliest stages is a key ingredient of efficient and effective target identification.

Key sources for this literature review include, among others, peer-reviewed research articles available through PubMed, information on clinical trials and outcomes, patents related to the disease and potential drug targets, and NIH grant awards. The challenge is effectively exploring and validating research hypotheses by finding and connecting all of the relevant bits of information. 

Several issues confront researchers in this process. First, the well known problems of language synonymy and polysemy are further exacerbated in Life Sciences, 
where the terminology is constantly growing as new discoveries are made, confounding manual efforts to curate ontologies. 

Second, exploring a drug target hypothesis typically requires connecting multiple pieces of information that describe different elements of a complex biological pathway. Since scientists are typically exploring novel hypotheses, there is no single source in the literature that provides the overall answer. Instead, the scientist must tediously find evidence spread across multiple sources for each component of the pathway.

Third, every hypothesis, or sub-component of that hypothesis, may have a variety of published results, some of which support the hypothesis, and some of which refute that hypothesis. Moreover, the researcher must consider the veracity of any single piece of evidence in support or refutation of a claim, which is usually a product of the prestige of the publication, the reputation of the authors and their institution, and the quality of the study or experimental methodology described in the evidence.

On the surface, all of these challenges sound like ideal candidates for an LLM solution. An LLM, however, provides only part of the solution.

Figure \ref{fig:gpt-medical} shows an example of using ChatGPT\footnote{https://chat.openai.com/} (as of Apr 2024) for medical research. The example question is about exploring a potentially multi-hop relationship between \emph{Rheumatoid Arthritis (RA)} and the inhibition of a particular kinase called \emph{IRAK4}. As shown in the figure, there are four main classes of problems with this purely LLM-based approach: the inability to control the search or ranking process, since the LLM is a black-box that is not grounded to a specific corpus where one might apply filtering or ranking criteria; the inability to validate results without cross-checking the references (which defeats the purpose of an efficient research solution); hallucinated references which undermine credibility of the approach; and the ``needle-in-the-haystack" problem as valid answers that are not popular in the training data are unlikely to be surfaced by the LLM.

As a result of these issues, the community has essentially migrated to a Retrieval Augmented Generation (RAG) approach for doing more precise and comprehensive search, where an LLM is combined with an Information Retrieval engine (search system) to produce answers that are grounded in a domain corpus, and are more up-to-date (beyond the training period of the LLM). There are several RAG based solutions that exist in the current marketplace, from general-purpose (web-based) search/answer generating engines like Perplexity\footnote{https://www.perplexity.ai/}, to domain-specific research solutions like Elicit\footnote{https://elicit.com/}. Since our focus area is on deep domain-specific research, we use Elicit as a baseline when doing Life Science research. 
Figure \ref{fig:elicit-qa} shows an answer to the same question with Elicit. Elicit produces an answer from the top 8 papers, and hence suffers from recall issues. Moreover, the answer is fairly shallow without providing a detailed causal understanding of the main biological linkages.

\begin{figure}[htb]
\centering
\includegraphics[width=0.8\columnwidth]{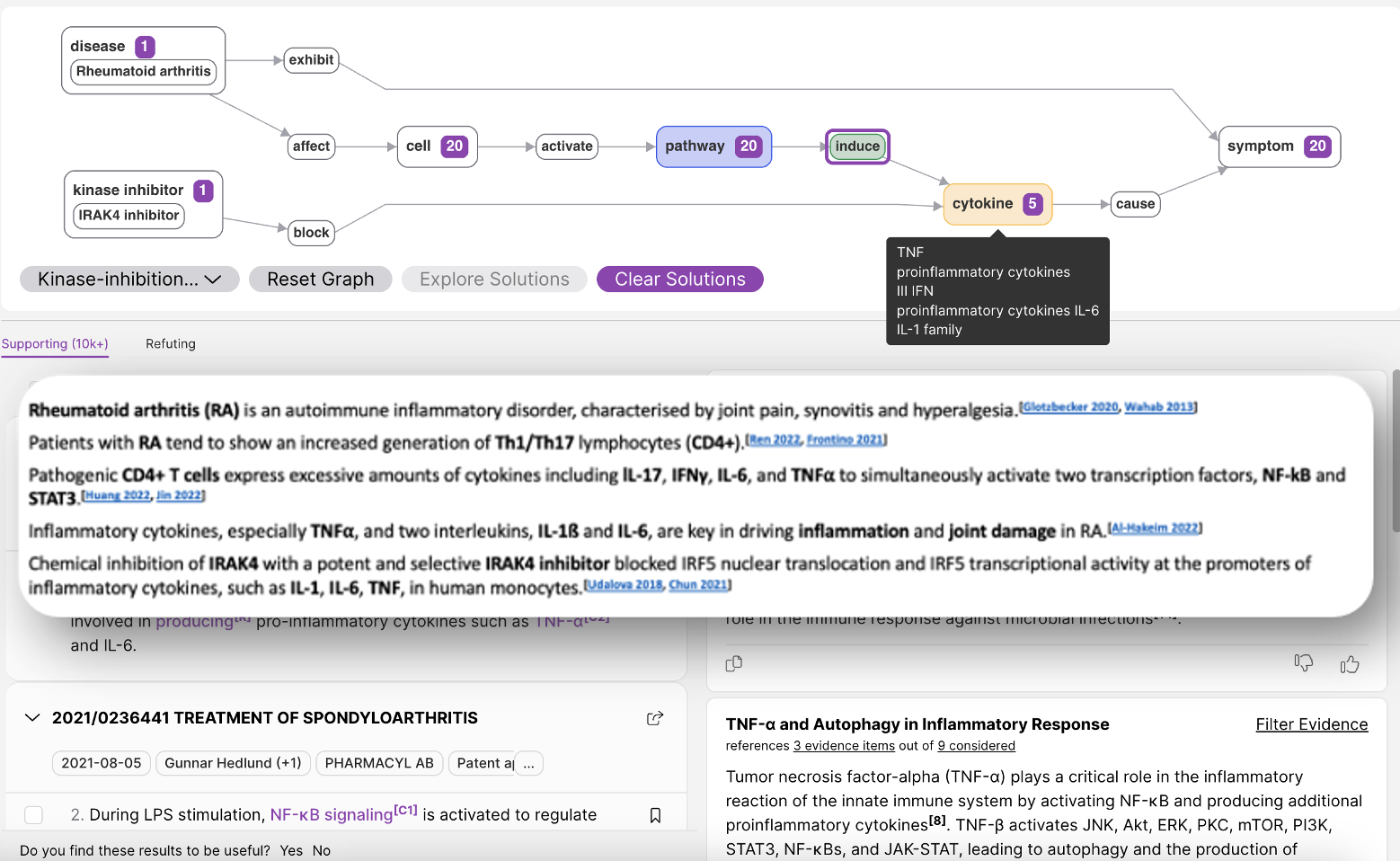}
\caption{\textbf{Cora's analysis for the IRAK4-RA question.} Cora extracts a detailed model linking RA and IRAK4 inhibitors based on a generalized research template, and produces a structured report with claims, evidence and citations}
\label{fig:cora-qa}
\end{figure}

Contrast this with the answer to the same question produced by our research assistant Cora, as shown in Figure \ref{fig:cora-qa}. Cora's approach is radically different in that it uses a general research template for connecting the dots between the two concepts of interest (RA and IRAK4 inhibitor), and then instantiates this template with specific bindings (answers) for concepts based on the inter-connected linkages. The research template is automatically induced from the data using our knowledge extraction algorithms and then further refined by a domain expert (the expert can directly specify a template as well). A single research template can be repurposed for multiple use-cases that involve the same kinds of concepts -- in this case, a link between any disease and kinase inhibitor, not just RA or IRAK4 inhibitors. The final answer produced by the system is based on the entire causal map and contains detailed evidence for each of the linkages with reliable citations.

\subsection{Macro-Economic Analysis: Multivariate Causal Inference}
\label{sec:macro-econ-use-case}

\begin{figure}[htb]
\centering
\includegraphics[width=0.8\columnwidth]{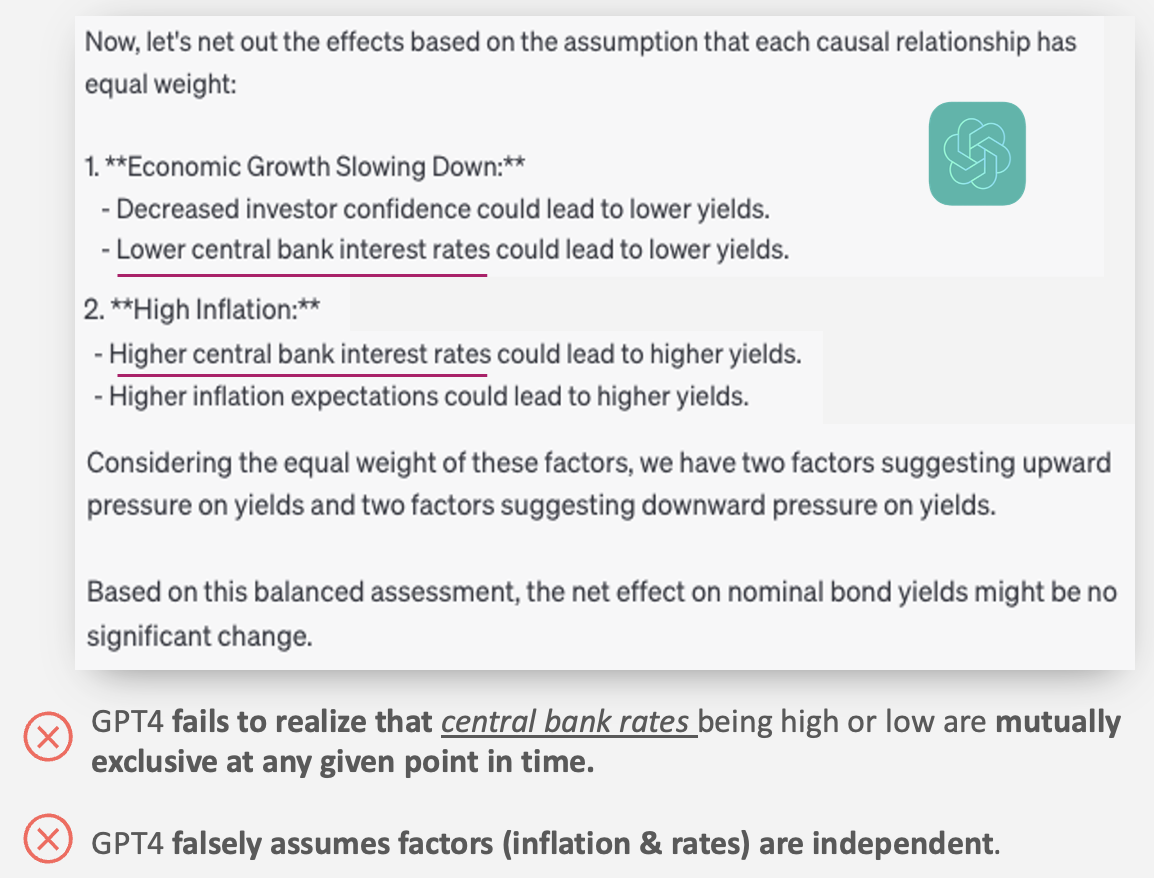}
\caption{\textbf{GPT4's response to the macro-economic question:} \emph{If economic growth is falling in an Emerging Market country, and the country is facing high inflation, what is the likely impact on nominal bond yields?}}
\label{fig:gpt4-yields}
\end{figure}

Consider the following macro-economic research problem: you are given a scenario that describes the state of a given economy, and the task is to understand the impact on a particular target concept. For example, \emph{``If growth is falling in an Emerging Market country, and the country is facing high inflation, what is the likely impact on nominal bond yields?"}. As can be seen, the scenario contains multiple economic factors (\emph{``falling economic growth"}, \emph{``high inflation"}) that are present in a given context (\emph{``Emerging Market country"}), and a good solution to this problem requires mapping out the relevant causal linkages between these factors and the target concept (\emph{``nominal bond yields"}), and performing causal inference to net out the influences.

We asked the above question directly to GPT4 and refined the prompt to get the model to analyze both upward and downward pressures on the target concept. The output is shown in Figure \ref{fig:gpt4-yields}. Apart from the fact that the answer does not include reliable references or data (for the reasons described in the previous section), the LLM also makes fundamental reasoning mistakes. In particular, it conflates mutually-exclusive conditions (that cannot be true at the same time), and asserts the possibility of both ``canceling each other out", which is logically invalid. Moreover, it assumes independence between highly inter-dependent factors, as shown in the example. 

While using a RAG-based GPT approach would help ground the results in a corpus and improve the reliability of the evidence, doing precise logical reasoning is still a capability that LLMs (even ones as powerful as GPT4) struggle with.

We adopt a neuro-symbolic approach to solving this problem. Instead of using a purely RAG-based approach for doing search and reasoning, we first use a multi-step graph search algorithm to identify relevant causal linkages in the text, and dynamically build a comprehensive causal map where each link is substantiated with evidence from the corpus. We then feed this causal map to a symbolic reasoning engine to propagate and reason over the causal influences considering the correlations and weights of various factors.

\begin{figure}[tbh]
\centering
\includegraphics[width=0.99\columnwidth]{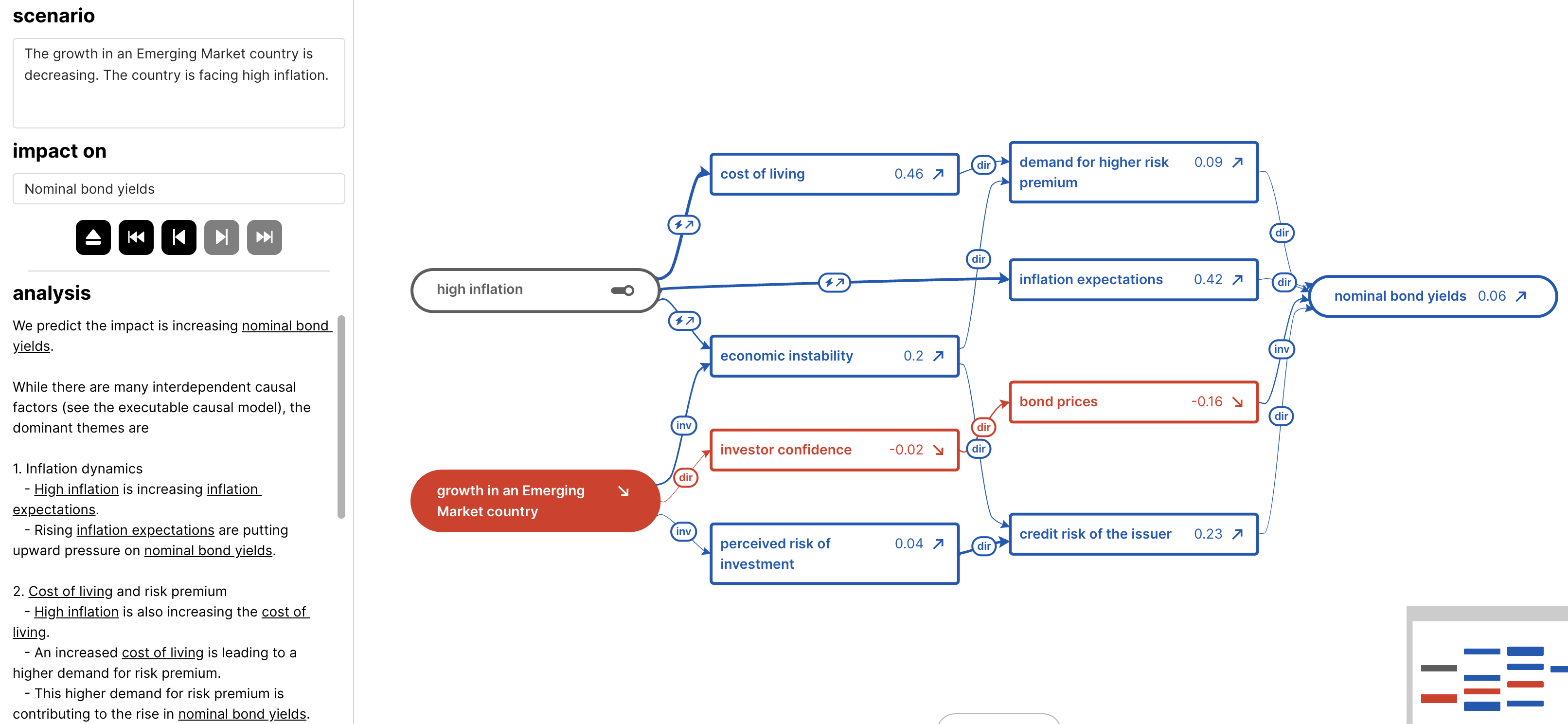}
\caption{\textbf{Cora's response to the question on nominal bond yields}. Cora extracts a scenario relevant causal map ``on-the-fly" from the corpus and does precise causal inference to compute the final result. Blue edges in the graph indicate upward pressure on the target node, while red edges indicate downward pressure. Similarly, the node color being blue or red depicts whether the quantity is increasing or decreasing respectively. The graph is fully interactive and the user can alter edge weights, add or remove nodes/edges and redo the causal inference on the fly.}
\label{fig:theoUI}
\end{figure}

Cora leverages these technologies to do causal inference over unstructured data. Figure \ref{fig:theoUI} shows the causal map extracted by Cora for the earlier question. Cora's answer and structured explanation (shown in the left of the figure) is generated from the causal map and describes the dominant themes at play, with the underlying causal chains.


The extracted causal map is a fully executable logical model, where the user can refine any part of the graph structure - e.g. drop edges, merge nodes, force the values of specific nodes (based on known facts or hypothetical scenarios), alter edge weights (based on domain knowledge) etc. - and then re-run inference to compute the effects of the changes. In this manner, Cora supports interactive precise counterfactual reasoning and \emph{What-if} analysis. As a next step, we are exploring connecting the causal map extracted from theory with real economic time-series data to make more informed statistical predictions.

\section{Neuro-Symbolic AI Platform}
\label{sec:platform}

In the previous section, we described challenges faced by LLM/RAG based approaches when tackling complex causal question answering problems that involve multiple factors and pieces of evidence and the Cora application we built to tackle these issues. In this section, we describe our AI platform used to build Cora, the analytics pipeline that constructs the semantic indices (KBs) from unstructured text, and highlight how symbolic reasoning is used to produce answers. 

\subsection{High-Level Architecture}
\label{sec:architecture}

\begin{figure}[tbh]
\centering
\includegraphics[width=0.99\columnwidth]{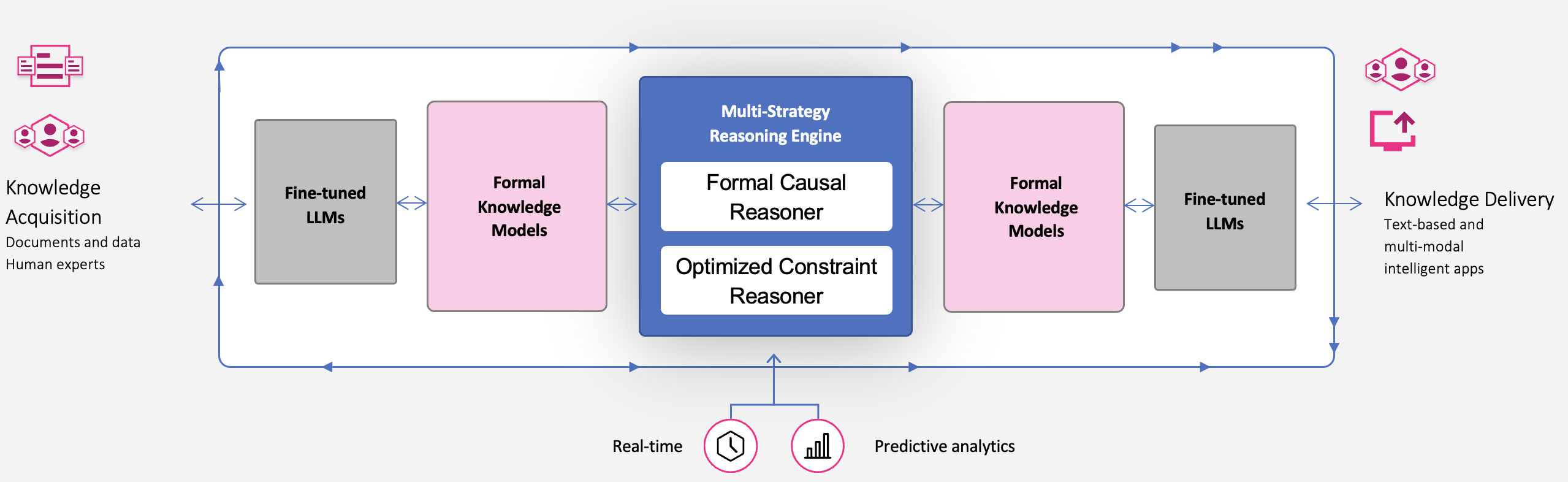}
\caption{EC's Neuro-symbolic AI Platform}
\label{fig:platform}
\end{figure}

As depicted in Figure \ref{fig:platform}, at the heart of the EC AI platform is a general-purpose symbolic multi-strategy reasoning engine that is based on Answer Set Programming \cite{asp}, and uses and builds on solvers such as Clingo \cite{clingo}. The reasoning engine performs the key function of logical reasoning including causal, deductive, abductive and non-monotonic inference as well as multi-objective constraint optimization based on given rules and facts. It supports interactive and incremental reasoning, involving the user to address knowledge gaps, resolve ambiguities, and make knowledge updates to the model on-the-fly, providing detailed explanations of the model's analysis.

In addition to these core interactive reasoning capabilities, the EC AI platform integrates with LLM-powered interfaces to facilitate knowledge acquisition and user interactions through fluid natural language interactions.

We support two kinds of knowledge acquisition -- semi-automated expert guided authoring for small-medium sized domain models, and fully-automated knowledge extraction from large domain corpora. The former uses our proprietary Knowledge Representation language known as \emph{Cogent}, and is the focus of the work described in \cite{chucarroll2024llms}. In this paper, we focus on our capabilities for the latter use-case.  

\begin{figure}[tbh]
\centering
\includegraphics[width=0.9\columnwidth]{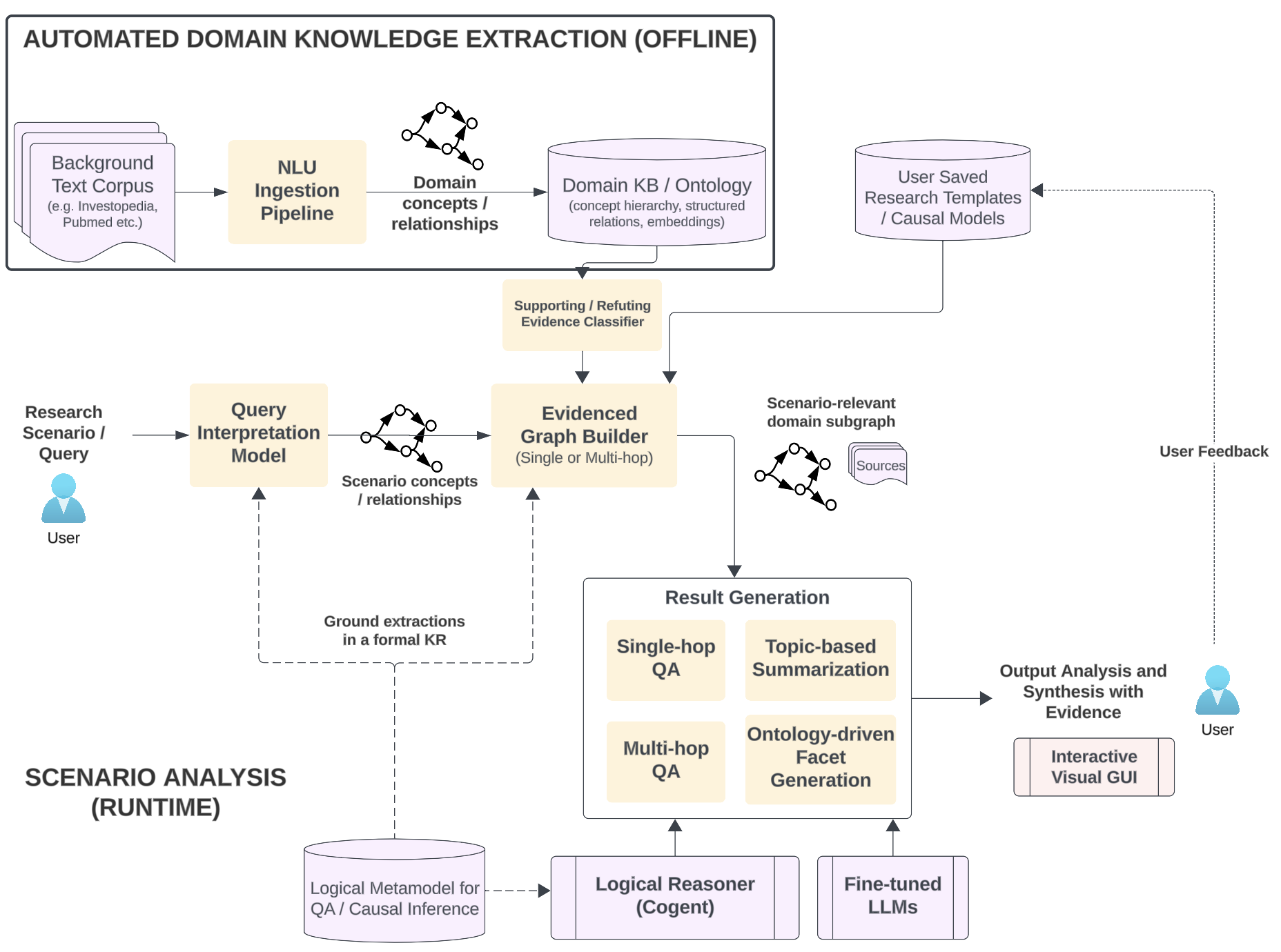}
\caption{Solution Architecture for Cora}
\label{fig:soln-arch}
\end{figure}

Figure \ref{fig:soln-arch} shows the solution architecture for Cora. There are two phases of the system - during the offline domain knowledge extraction phase, the system ingests a text corpus using EC’s Natural Language Understanding (NLU) pipeline (more on this in the next section), extracts domain concepts and relationships in the text, and stores the resultant structured information along with text embeddings (vectors) for concepts and passages in a semantic index (“Domain KB”). 

At runtime (i.e. online phase), the system is given a user scenario, and it uses a Query Interpretation Model (a fine-tuned LLM) to process the input and pull out salient concepts and relationships in the scenario question. This information is passed to the Evidenced Graph Builder, which uses an iterative, multi-step graph search algorithm to find relevant relationships and chains from the Domain KB that are applicable to the input scenario (alternately, it might retrieve an applicable pre-saved research template when relevant). The algorithm fleshes out a relationship/causal graph connecting input concepts in the scenario to the target/query concept, where each link is sourced from the theory texts.

The Evidenced Graph Builder is also fed the QA/Inference Meta-model as one of its inputs. This is an abstract logical meta-model for doing causal inference and question answering, and provides the logical scaffolding (or grounding) for the extracted concepts and relationships. It is specified using our proprietary Cogent language. The meta-model draws on an established causal reasoning framework, Qualitative Process Theory, as well as its recent applications to knowledge graph extraction, as a starting point to formalize concepts such as Quantities, States, and the causal influences that propagate between them \cite{forbus1984,forbus2019, friedman2022unstructured}.

The output of the Evidenced Graph Builder is an instance of this meta-model that is specialized to the scenario concepts and the extracted causal linkages. This model is then executed using the Cogent Reasoning Engine (RE) to derive inferences based on the specific connections in the map.

The Result Generation module uses the Cogent RE along with fine-tuned LLMs to provide various functionalities, ranging from single/multi-hop QA to topic-based summarization and facet generation. The latter two features are beyond the scope of this paper. 

Finally, the system allows users to save the analyzed and reasoned-over knowledge graph results, which can be reused for future scenario analysis.

\subsection{Knowledge Extraction using Statistical Models}
\label{sec:OntInduction}

The NLU Ingestion Pipeline is used to process a text corpus and extract domain knowledge. At EC, we have designed a general purpose Meaning Representation Schema to capture knowledge. The schema is centered around the notion of \textbf{contextual Relationships or Events}, where each relationship is characterized by its subject and object concepts, along with \emph{qualifiers} that specify contextual information about \emph{time, space, manner, purpose} etc. Additionally, concepts and relationships are arranged in a hierarchy to support taxonomic reasoning. Our schema is inspired by KR formalisms such as AMR \cite{AMR}, FrameNet \cite{framenet}, PropBank \cite{propbank}, etc., but is designed to be leaner and more concise to aid generality. 

\begin{figure}[tbh]
\centering
\includegraphics[width=0.6\columnwidth]{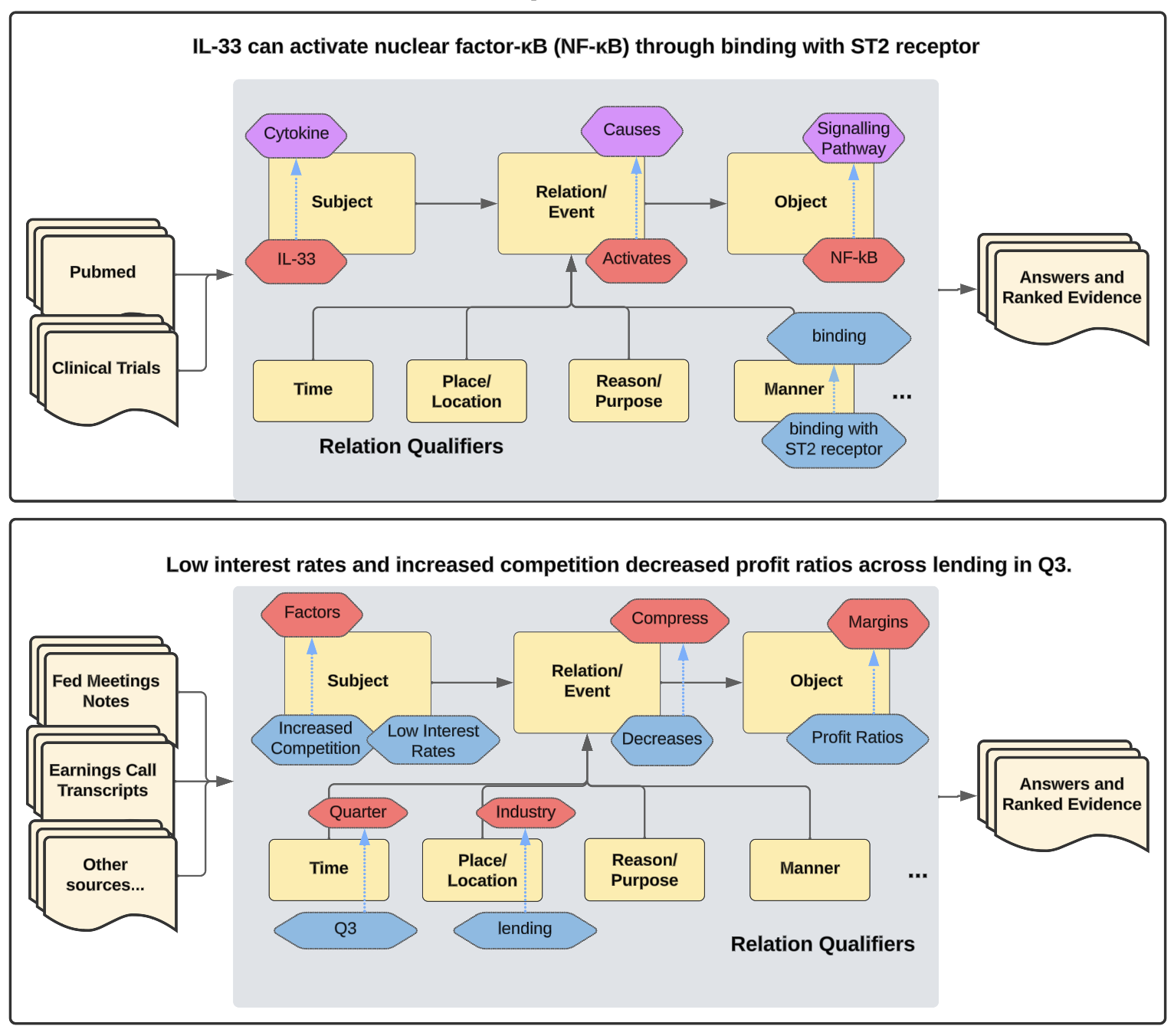}
\caption{Rich Conceptual Event Structure}
\label{fig:eventStructure}
\end{figure}

Figure \ref{fig:eventStructure} shows examples of events extracted from sentences in two different domains - medical and finance. In the medical example, given the sentence shown, we extract the relationship ``activates" between ``IL-33" and ``NF-Kb", along with the qualifier \emph{manner}: ``binding with ST2 receptor". Moreover, we also extract type information for the concepts taking the context into account when disambiguating its meaning -- in this case, ``IL-33" is an instance of ``Cytokine", while ``NF-Kb" is an instance of a ``Signalling Pathway". This rich event structure lets us answer questions such as: \emph{``Which cytokine activated a signalling pathway and how was it done?"}. The same holds for the financial example, since the schema is domain independent.


Our process of Ontology Induction involves extracting rich event (relational) structures as shown in the figure, and the underlying type hierarchy from the text, and we do this in a fully unsupervised manner - i.e. with no manual training data. Additionally, we perform \emph{Entity Linking} (similar to \cite{entity-linking}), in order to link the induced concepts from the text with entities in external knowledge bases and ontologies.

\begin{figure}[tbh]
\centering
\includegraphics[width=0.9\columnwidth]{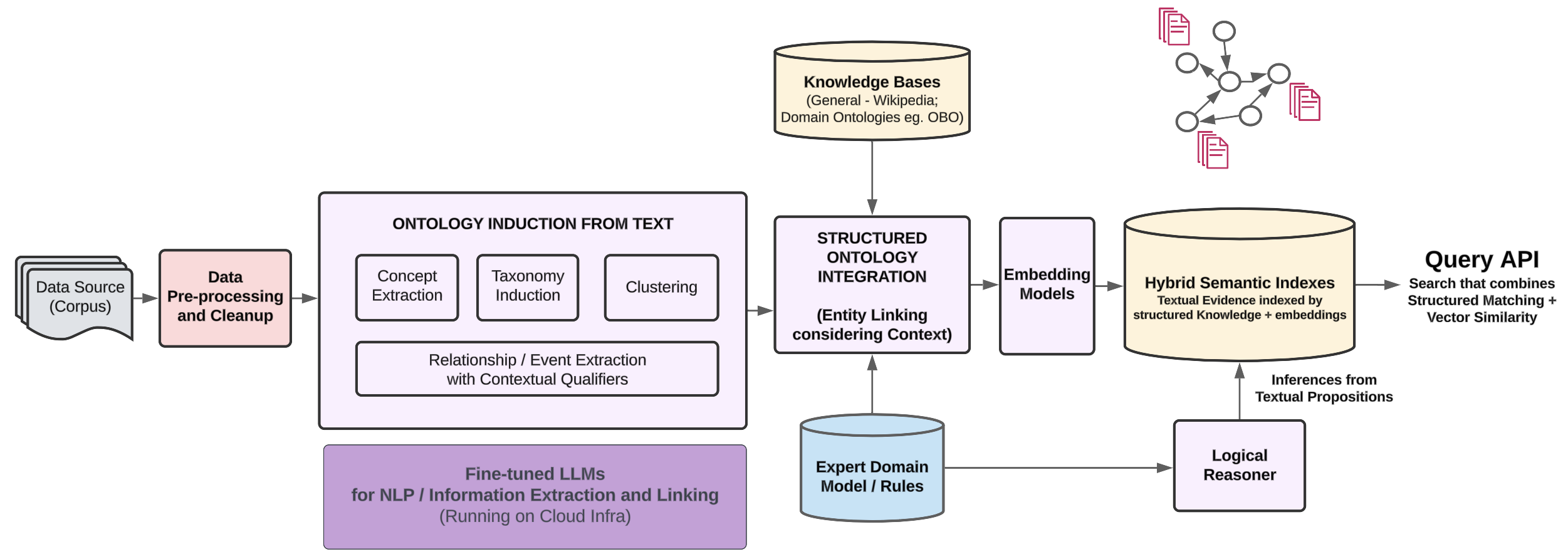}
\caption{NLU Ingestion Pipeline}
\label{fig:nluIngestion}
\end{figure}

For this problem, we have developed a transformer-based architecture for Linking and Understanding Mentions using Encoder Networks, LUMEN, that we use for concept typing, entity linking and relationship extraction, along with a fully automated domain-adaptation process that leverages our own fine-tuned LLMs to generate synthetic training data. The framework uses SLMs (Small Language Models such as Gemma-2B \cite{gemmateam2024gemma}) for increased throughput (sub-200ms latency per passage) without compromising on quality, by using high quality training data and novel contrastive loss functions. Details of this framework will be provided in a forthcoming technical publication. 

\subsection{Multi-hop QA and Explanations using Symbolic Reasoning}

\subsubsection{Cogent: KR Language and Meta-Model}

Cogent is EC’s proprietary Knowledge Representation language. It can be used to formally define conceptual theories in a form of structured English. Its underlying formalism is based on Answer Set Programming and supports term definitions, rules, (hard/soft) constraints and objective functions. Additional details of Cogent are in \cite{chucarroll2024llms}. 

As mentioned earlier, we have defined a general meta-model for causal inference in Cogent, which provides the logical grounding for terms and relationships, and facilitates reasoning via the Cogent-RE. 

\begin{figure}[tbh]
\centering
\includegraphics[width=0.9\columnwidth]{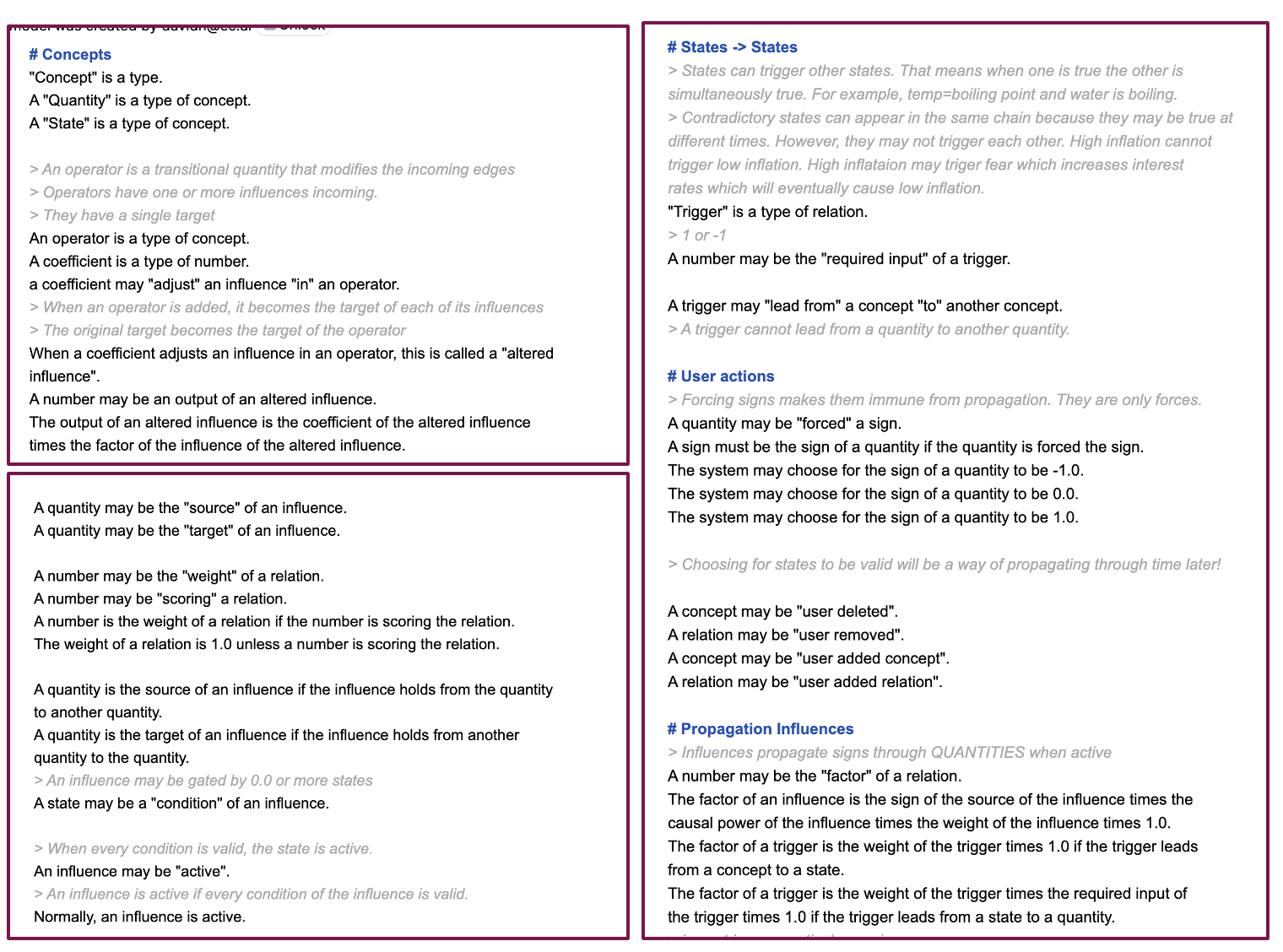}
\caption{Cogent Meta-Model for Causal Inference}
\label{fig:cogentMM}
\end{figure}

Figure \ref{fig:cogentMM} shows snippets of our Cogent meta-model for Causal Inference, based on Qualitative Process Theory (QPT). 

QPT provides a robust framework for understanding the dynamics of continuous systems through the propagation of qualitative values (e.g., increasing, decreasing, high, low...), rather than relying solely on numerical data. This makes it applicable across diverse fields where under-specified values are prevalent such as economics, medicine, geopolitics, and cybersecurity.

Under QPT, quantities are causally influenced by \emph{processes}, and the effects of that influence propagates between quantities. As an example, in a heat flow process, the heat transfers from a hot to a cold object. As the cold object heats up, that may cause subsequent changes (e.g., maybe it becomes more malleable) \cite{forbus1984,forbus2019}. Additionally, \cite{friedman2022unstructured} took inspiration from QPT's quantity-to-quantity propagation in their work by annotating causal models in natural language.

Like them, our representation draws inspiration from QPT's influence mechanism between quantities, but we further expand our approach to include the notion of ``States" and the ``Triggers" causal relationship.

In economics, \texttt{Quantities} include variables like \emph{GDP} and \emph{interest rates}, while \texttt{States} might describe those quantities at a specific value such as \emph{high inflation} or inciting events like the \emph{imposition of tariffs}. Extending to medicine, \texttt{Quantities} could encompass fluctuating metrics like \emph{blood pressure or cholesterol levels}, with \texttt{States} representing medical conditions such as \emph{diabetes} or \emph{stages of cancer remission}. Similarly, in cybersecurity, \texttt{Quantities} include metrics like the \emph{number of system intrusions} or \emph{data transfer rates}, and \texttt{States} could refer to the \emph{security status of systems} or the \emph{occurrence of breaches}. This framework allows for a nuanced analysis of how various factors interact within and across these fields, providing insights into how changes in one area can influence outcomes in another, thereby offering a comprehensive view of causal dynamics in complex environments.

In our meta-model, \texttt{Influences} describe the causal relationship between two quantities, which can be either direct or inverse, indicating how one quantity affects another. For instance, in medicine, an \emph{increase in medication dosage} might influence the \emph{reduction of symptom severity}. In geopolitics, \emph{a rise in military expenditure} might inversely influence the \emph{economic stability of a country}.

\texttt{Triggers}, however, define the causal relationships between states or between a state and a quantity, highlighting how certain states can act as tipping points, initiating changes in other states or quantities. For example, in cybersecurity, the \emph{detection of a new malware type} (a \texttt{State}) might trigger an \emph{increase in security protocol updates} (a \texttt{Quantity}).

This structured approach provides a deeper understanding of the mechanisms driving changes within dynamic systems and allows us to use Cogent to interactively reason about what-if scenarios, providing valuable insights into complex causal interactions in various domains.

\subsubsection{Evidenced Graph Building and Symbolic Reasoning}

As mentioned in Section \ref{sec:architecture}, the Evidenced Graph Builder's goal is to find multi-hop relational and causal chains connecting input concepts in the scenario to the query or target concepts. This problem can be cast as a graph search or path-finding problem, where each link in the path is a causal relationship, the source nodes are the input concepts, and the target node is the query concept. Here, the aim is not to find the shortest causal path from source to target, but instead to find all dominant and contextually relevant paths linking the source and target nodes based on the domain knowledge. Longer paths that reveal more details are preferred as they help build a better causal understanding of the scenario. 

Our solution is based on the A* Search algorithm \cite{a-star}, which runs a forward-backward search routine (i.e. forward from the source nodes, backward from the target) and uses an LLM (in particular, its intrinsic world knowledge) as the search heuristic to estimate which paths are likely to connect up from both sides. The algorithm also has plugin points for automatically inserting relevant user knowledge from prior saved causal maps. 

As part of producing the map, the builder maps the specific relations (predicates) in the event structures stored in the Domain KB (described in Section \ref{sec:OntInduction}) to the higher-order relationships such as ``influence" and ``trigger" in the Cogent Meta-model. 

The final knowledge graph produced by the builder, which is an instantiation of the meta-model, is an executable logic program that is fed to the Cogent RE. The result of reasoning is fed to an LLM to produce the final answer and explanation from the detailed logical proof traces.

\section{Initial Evaluation}




\label{sec:evalMHQA}



Our goal is to answer the following question: Given a complex research query that involves multiple concepts and relationships, how accurately and comprehensively can an AI system provide detailed answers and explanations that consider relevant intermediate links and include citations for supporting and/or refuting evidence?

We report results of an initial evaluation conducted in the medical domain.

\subsection{Medical QA Eval}

We collected two sets of 25 queries each based on real questions from experts in the medical research domain.
\begin{itemize}
    \item \emph{Representative Queries}: The first set is composed of queries that are representative of the types of queries experts could ask of a medical QA system. This set kept the queries as they were asked by the experts. (see Appendix \ref{sec:repr_queries} for the list of the queries.) 
    \item \emph{Multi-hop Queries}: The second set was specifically selected to evaluate ``multi-hop" question answering. A satisfactory answer to such queries would need to outline complex causal relations involving multiple intermediary causal factors. In order to avoid confounds due to surface form variations (which is already addressed by the first data set) and to facilitate the evaluation of answer complexity, the queries were reformatted to follow the format ``How does X impact Y?". (see Appendix \ref{sec:mh_queries} for the list of the queries.) 
\end{itemize}

The queries were evaluated using four systems:
\begin{itemize}
\item \emph{GPT4-Turbo}\footnote{https://platform.openai.com/playground/chat?models=gpt-4-turbo}: a state-of-the-art LLM
\item \emph{Perplexity}\footnote{https://www.perplexity.ai/}: RAG using web-search
\item \emph{Elicit}\footnote{https://elicit.com/}: RAG using Semantic Scholar for doing scientific research
\item \emph{Cora}: Our solution.\footnote{Although our systems allows a user to further refine an answer, we here evaluate the direct answer from Cora to a user's question without any further user interactions.}
\end{itemize}

All four systems mentioned above were run on the questions and we asked each system to produce an answer with supporting/refuting evidence and cited sources.

We focus on evaluation on intrinsic task of \emph{verification} and \emph{relevance} of answers as opposed to direct comparison with what would be a necessarily incomplete gold standard answer as each question could have multiple equally correct answers. 
Domain experts manually reviewed each of the systems' results. They were asked to check that the generated answer/explanation justifies its claims, is both accurate and relevant, and that the sources it cites actually exist. 

We designed the following metrics to assess whether or not a user can verify the claims presented and whether or not those are relevant:


\begin{enumerate}
    \item \textbf{Claim Density}: Average number of claims per answer. \emph{A measure of the quantity of information provided}.
    \item \textbf{Citation Density}: Average number of real citations per claim. \emph{A measure of the amount of verification options}.
    \item \textbf{Source Hallucination Rate}: Percentage of citations that are not valid (real) and scholarly sources. \emph{A measure of system hallucination}.
    \item \textbf{Citation Rate}: Percentage of claims in the answer that are accompanied by real citations. \emph{A measure of verifiability of the overall answer}.
    \item \textbf{Justification Rate}: Percentage of claims that are a correct paraphrase of a real citation. \emph{A measure of interpretation quality}. Claims with non-existent sources are not justified as they are unverifiable. Since checking this requires manual effort, we imposed a max time-limit of 5 minutes on the domain expert to verify each claim.
    \item \textbf{Relevance Rate}. Percentage of claims that are justified and relevant to answering the question. \emph{A measure of relevance and quality of answer}.
\end{enumerate}

Results are shown in Table \ref{tab:shqa} and \ref{tab:mhqa1}. Note that the metrics from 4-6 get progressively stricter, as a justified claim must also be cited, and a relevant claim must also be justified.

\begin{table}[htb]
\small
    \centering
    \resizebox{\columnwidth}{!}{
    \begin{tabular}{|c|c|c|c|c|c|c|} \hline
        \textbf{System}     & \textbf{Claim Density} & \textbf{Citation Density} & \textbf{Source Hallucination Rate} & \textbf{Citation Rate} & \textbf{Justification Rate} & \textbf{Relevance Rate} \\ \hline
         GPT4-Turbo          & 3.52                   & 0.63                      & 42.86\%                            & 47.73\%                & 25.00\%                     & 23.86\%  \\ \hline
         Perplexity          & 3.60                   & 0.38                      & \textbf{0.00\%}                    & 22.22\%                & 21.11\%                     & 10.00\% \\  \hline
         Elicit              & 3.96                   & \textbf{1.53}             & 3.82\%                             & 95.96\%                & 82.83\%                     & 69.70\% \\ \hline
         Cora                & \textbf{5.04}          & 1.26                      & \textbf{0.00\%}                    & \textbf{100.00\%}      & \textbf{93.65\%}            & \textbf{84.92\%}  \\ \hline
    \end{tabular}}
    \caption{Representative Queries. Results with Evidence and cited Sources.}
    \label{tab:shqa}
\end{table}

\begin{table}[htb]
    \centering
    \resizebox{\columnwidth}{!}{
    \begin{tabular}{|c|c|c|c|c|c|c|} 
    \hline
    \textbf{System} & \textbf{Claim Density} & \textbf{Citation Density} & \textbf{Source Hallucination Rate} & \textbf{Citation Rate} & \textbf{Justification Rate} & \textbf{Relevance Rate} \\ \hline
     GPT4-Turbo  & 4.16  {\small(+0.64)} & 1.01 {\small(+0.38)}  & 31.4\% {\small(-11.46)}  & 64.42\% {\small(+16.69)} & 27.88\% {\small(+2.88)} & 22.12\% {\small(-1.74)}  \\ \hline
     Perplexity  & 4.76 {\small(+1.16)} & 0.59 {\small(+0.21)}  & 0.01\% {\small(+0.01)} & 32.77\% {\small(+10.55)} & 17.65\% {\small(-3.46)} & 11.76\% {\small(+1.76)} \\  \hline
     Elicit   & 5.00 {\small(+1.04)}     & \textbf{1.36} {\small(-0.17)} & 0.01\% {\small(-3.81)} & 98.40\% {\small(+2.44)} & 86.40\% {\small(+3.57)} & 60.80\% {\small(-8.90)} \\ \hline
     Cora     & \textbf{5.36} {\small(+0.32)} & 1.14 {\small(-0.12)} & \textbf{0.00\%} {\small(+0.00)}  & \textbf{98.51\%} {\small(-1.49)}  & \textbf{90.30\%} {\small(-3.35)} & \textbf{86.57\%} {\small(+1.65)} \\ \hline
    \end{tabular}}
    \caption{Multi-hop Queries. Results with Evidence and cited Sources. Values in parentheses indicate the difference with Representative Queries results of Table \ref{tab:shqa}}
    \label{tab:mhqa1}
\end{table}

In addition, for the answers to the Multi-hop Queries dataset, we were interested in quantifying the complexity of the answers generated. Given that the queries are of the type ``What is the impact of X on Y?", a domain expert recorded for each answer:
\begin{enumerate}
    \item \textbf{Maximum Number of Hops}: Maximum number of hops (relations) tying the source (X) to the target (Y) in a reasoning chain. (e.g. 'X causes A1 causes A2 causes Y' would be a 3-hop reasoning chain since it is composed of 3 causal relations). This an indication of the \emph{depth of the answer}.
    \item \textbf{Number of Concepts}: Number of concepts presented in the answer that are directly relevant to the explaining the mechanism. This is an indication of \emph{coverage of the answer}.
\end{enumerate}

Results are shown in Table \ref{tab:mhqa2}.

\begin{table}[htb]
    \small
    \centering
    \begin{tabular}{|c|c|c|} 
    \hline
    \textbf{System}  & \textbf{Maximum Number of Hops} & \textbf{Number of Concepts} \\ \hline
     GPT4-Turbo  & 2.5 $\pm$2.1 & 5.1 $\pm$3.1  \\ \hline
     Perplexity   & 1.5 $\pm$1.2 & 4.0 $\pm$3.3  \\  \hline
     Elicit    & 0.8 $\pm$0.6 &  3.3 $\pm$3.2    \\ \hline
     Cora     & 2.1 $\pm$0.7 &  7.5 $\pm$2.4    \\ \hline
    \end{tabular}
    \caption{Multi-hop Queries. Answer Complexity Analysis.}
    \label{tab:mhqa2}
\end{table}

\subsection{Discussion}
\subsubsection{Results on Representative Queries data}
Across the four systems evaluated for the queries, we find that Cora and Perplexity are the only two systems that reliably cite articles that exist. Sourcing claims in real evidence is crucial in the medical domain, as there is minimal room for error and misinformation with experts facing decisions that are high-stakes. It is particularly worth noting the \emph{Source Hallucination Rate} of GPT-4 Turbo in Table \ref{tab:shqa} with almost every other article being hallucinated. Even though both Cora and Perplexity cite real articles 100\% of the time, Perplexity only cites a few articles for a few of its claims, as evidenced by the low \emph{Citation Density} and \emph{Citation Rate} values.

For claims in the tools' answers that are derived from valid sources, it is important to know if the information as it is represented in the answer is justified by the information in the source. This is measured by the \emph{Justification Rate} metric. These metrics together make it evident that Cora provides the most reliable and verifiable claims. This is attributed to Cora's precise and granular evidence-based answer generation algorithm. Additionally, it is important that a researcher is able to quickly verify the consistency across the answer and the source, something they can do easily in Cora where relevant snippets (highlighted paragraphs) from the paper are directly evidenced. All the other tools require the expert to read the entire paper to find the precise evidence.

\emph{Claim Density} measures the quantity of information presented in the answer, and Cora provides the most comprehensive answers across the compared systems. Finally, the metric that evaluates the usefulness of an answer to a researcher/expert is the \emph{Relevance Rate}, which is obtained by considering how many of the total justified claims of a given answer are labeled as ``relevant" by an expert. The measure of relevance here was lenient as relevance can be subjective depending on the expert. Results show that Cora provides the most relevant and comprehensive answers that are backed by real evidence and represents the cited evidence well compared to the other tools. 

\subsubsection{Results on Multi-hop Queries data}
Results for the Multi-hop Queries are shown in Table \ref{tab:mhqa1}.
All the points discussed above still hold in the case of multi-hop queries.
GPT-4 Turbo and Perplexity, systems that had the lowest performances in the Representative Queries, display similar poor results on multi-hop queries. Crucially, looking at the final relevance rate, there is no drop in quality of answer in Cora while Elicit's final relevance score on the other hands drops by almost nine points.

The answer complexity analysis shown in Table \ref{tab:mhqa2} add another dimension to the results. A pure LLM solution such as GPT-4 Turbo generates answers with a relatively high number of concepts (good coverage) and with the longest reasoning chains of more than 2 hops on average (good depth). However, as shown in Table \ref{tab:mhqa1}, most of its claims are not justified and irrelevant. Conversely, Elicit, a system tailored for research, has a higher rate of justification and relevance but its answers contain fewer concepts (low coverage) and with fewer hops (lack of depth).
Cora's answers combine both a high coverage and depth with justified and relevant claims.

Cora distinguishes itself by the stability of its performances as well as the depth and coverage of its answers. This is due to EC's implementation that integrates LLMs with symbolic causal models into the answer generation process. This allows for the iterative exploration, capture, verification and refinement of relevant knowledge extracted from unstructured domain data into an integrated causal graph. This graph in turn provides a solid semantic scaffold to guide the text generation process.

\section{Conclusions}

The past two years have seen unprecedented excitement and investment in AI. Effectively leveraging LLMs and Generative AI is becoming a business imperative across every major industry, where business leaders are motivated by both seeking competitive advantages with more automation and intelligence, and fear of being left behind if they fail to effectively leverage AI. For modern AI to deliver on these enormous expectations, solutions must meet a wide variety of critical requirements, not the least of which is reliable and accurate answers users can trust to make critical business decisions. 


At Elemental Cognition our mission is to deliver on the promise of AI with technology that provides more accurate, relevant, and verifiable answers and solutions to complex business problems. To this effect, we have developed a neuro-symbolic AI platform that combines two powerful complementary technologies - statistical language understanding machines (LLMs) and symbolic reasoning engines. The former is used to extract, formalize and translate knowledge from text, while the latter is used to precisely analyze, reason and explain answers. 

We have described our approach and architecture in detail and presented experimental results validating its performance. Our results show that our solution delivers best in class performance at answering complex research questions requiring multi-step exploration, reasoning and justification over a knowledge-intensive domain.


\section{Acknowledgements}
We would like to thank the following people for their contributions to the EC AI Platform and the Cora product: Mike Barborak, Cliff Kamppari-Miller, Natalie Dunn, Scott Mandrell, Josh Griffith, Elia Lake, Blake Harris, Jeremy Martin, Ross Bogel, Asia Gagnon, John Mumm, Tomas Silva, Shirin Saleem.

\bibliography{bibliography}

\section{Appendix}
\subsection{Representative queries data set}
\label{sec:repr_queries}
The following queries were used for representative query evaluation:
\begin{enumerate}
\item What is the function of ZAP-70?
\item Is CD163 activity increased in multiple sclerosis?
\item What role does the innate immune system have during rheumatoid arthritis?
\item Flexibility of E3 ligase Cbl domains
\item What do LALA mutations do for antibodies?
\item What are some cytotoxic payloads when using ADC drugs?
\item What is the target of AR-A014418?
\item what are synonyms for CHEK2?
\item Why did LOXL-2 inhibitors fail clinical  trials?
\item TLR agonist for treatment of bladder cancer
\item What is the role of NLRP in inflammation?
\item Does the epigenetic dysregulation of CTRB1 contribute to the development of diabetes?
\item What are the molecular pathways involved in the tumor environment of breast cancer?
\item How is the loss of RHO linked to podocyte function?
\item Which SNPs are associated with PARK variants?
\item What cytokines and chemokines are secreted by colorectal cancer cells?
\item Does BACH1 promote angiogenesis?
\item What are some markers for liver stellate cell damage?
\item Which diseases are affected by THOC2?
\item Does clodronate deplete macrophages?
\item At what sites is CCR6 mostly expressed in?
\item How does the transcriptional repression of p53 happen?
\item What are some pharmacovigilance concerns for IRAK4 inhibitors?
\item Is there evidence for lung cancer being resistant to treatment due to EGFR mutations?

\item How do macrophages play a dual role in brain cancer?
\end{enumerate}

\subsection{Multi-hop Queries data set}
\label{sec:mh_queries}
The following queries were used for multi-hop query evaluation:
\begin{enumerate}
\item How does FOXO impact cell death?
\item How does Kidney disease impact LRKK2?
\item How does neutrophils impact multiple sclerosis?
\item How does GLPR agonist impact diabetes?
\item How does IL-2 expression impact inflammatory disease progression?
\item How does epigenetic dysregulation of neurotrophins impact AD risk?
\item How does PI3K pathway impact TME in prostate cancer?
\item How does IGF-1 overexpression impact acne?
\item How does TRIM8 impact hypoxia?
\item How does PFN1 impact glucose regulation?
\item How does ischemic stroke occurring impact NLRP3 activation?
\item How does exercise impact mitochondrial metabolism?
\item How does YAP signaling impact residual cancer?
\item How does FLT3L impact immune cell production?
\item How does APOE expression impact amyloid beta plaques?
\item How does estrogen deficiency impact osteogenesis?
\item How does AKT1 impact pulmonary hypertension?
\item How does CD24 overexpression impact cell migration?
\item How does CD74 impact Treg activity?
\item How does Epstein-Barr Virus infection impact epithelial cells?
\item How does Argininosuccinate synthase impact tumor?
\item How does SOD1 being mutated impact amyotrophic lateral sclerosis?
\item How does LncRNAs impact phagocytosis?
\item How does psilocybin impact depression?
\item How does Ischemic reperfusion injury impact HIF-1 expression?
\end{enumerate}

\subsection{Multi-hop answer comparison}
\label{sec:mhans_comp}

To illustrate the differences between Cora and Elicit, we provide here a qualitative comparison of the answers generated by both systems for the same multi-hop query ``How does epigenetic dysregulation of neurotrophins impact AD risk?" shown in Table \ref{tab:mhans_comp}.

\begin{table}[htb]
    \small
    \centering
    \begin{tabular}{|c p{13cm}|} 
    \hline
    & \textbf{Query}: How does epigenetic dysregulation of neurotrophins impact AD risk?\\ 
    \hline
    & \textbf{Cora Answer}\\ 
    C1 & Epigenetic dysregulation of neurotrophins, particularly through increased promoter CpG methylation, leads to a down-regulation of brain-derived neurotrophic factor (BDNF) expression, which is associated with decreased synthesis of BDNF in neurons [1][2].\\ 
    C2 & This dysregulation is linked to reduced neuroprotection and has been implicated in the emergence of various neurological and psychiatric diseases [3][4].\\ 
    
    C3 & BDNF is crucial for cognitive function, promoting neurogenesis, neuronal survival, axonal and dendritic growth, synaptic plasticity, and maintenance in the central nervous system [5].\\ 
    
    C4 & Consequently, decreased levels of BDNF result in the reduction of synaptic proteins, leading to synaptic connection loss, impaired neuronal survival, and adaptability, which are critical factors in neurodegeneration [6].\\ 
    
    C5 & This pathway of dysregulation contributes significantly to Alzheimer's disease (AD) neuropathology, particularly through the increase of hippocampal neurofibrillary tangles (NFTs) and plaques (NPs), which are strongly correlated with cognitive decline and are hallmark features of AD [7][8].\\

    C6 & Thus, epigenetic dysregulation of neurotrophins, by decreasing BDNF expression and signaling, directly impacts AD risk by promoting mechanisms that lead to neuronal and synaptic degeneration.\\
    \hline
    & \textbf{Elicit Answer}\\ 
    E1 & Epigenetic dysregulation, particularly in the form of DNA hypermethylation and histone deacetylation, has been implicated in the pathophysiology of Alzheimer's disease (AD) (Chouliaras 2010, Sanchez-Mut 2015, Millan 2014).\\ 
    
    E2 & This dysregulation can lead to a repressed chromatin state and reduced plasticity, potentially contributing to the onset and progression of AD (Sanchez-Mut 2015).\\ 
    
    E3 & The dysregulation of neurotrophins, a family of proteins that support the survival and growth of neurons, is a key aspect of this epigenetic dysregulation (Zusso 2018). In particular, the dysregulation of brain-derived neurotrophic factor (BDNF) has been linked to AD risk (Nativio 2018).\\ 
    
    E4 & This suggests that targeting the epigenetic dysregulation of neurotrophins, including BDNF, could be a potential therapeutic strategy for AD (Lardenoije 2015, Qureshi 2011, Daniilidou 2011).\\ 
    \hline
    \end{tabular}
    \caption{Comparison of Elicit and Cora answers to the query ``How does epigenetic dysregulation of neurotrophins impact AD risk?"}
    \label{tab:mhans_comp}
\end{table}

At first glance and for a non-expert, Elicit provides an answer that seems to span a variety of concepts (coverage) and be composed of claims that are all accompanied by citation (verifiability). Inspection by a domain expert, however, reveals the following:

\begin{itemize}
\item \textbf{Concept granularity mismatch}: Whereas the question asked about ``epigenetic dysregulation of neurotrophins" specifically, Elicit's answer opens (E1) by mentioning the existence of a relation between the generic phenomenon of epigenetic dysregulation and AD. The clause introduced by the adverb ``particularly" does not provide the specification it promises since it only describe generic processes through which epigenetic dysregulation can occur. This concept granularity mismatch introduced in E1 is carried over to E2 (``This dysregulation...").

This contrasts with Cora's opening statement (C1), contrast made all the more visible that similar grammatical constructions are used. Here the right concept is targeted and the clause introduced by the adverb ``particularly" further specifies the process by which the dysregulation of interest takes place.

\item \textbf{Unjustified claims}: After examining the references cited in E2, E3, and E4, a domain expert concluded that a those do not back the associated claims. Conversely, all the evidence that were cited by Cora were deemed to support their associated claim

\item \textbf{Vagueness of effects}: Most statements in the Elicit answer are vague with respect to the nature of the net causal impact of factors. E1 mentions an ``implication" without going further. E3 simply indicates ``a link to AD risk" without specifying how the nature of this link.
Most of Cora's statements clearly qualifies the nature of the effects: ``decrease synthesis of BDNF", ``impaired neural survival", ``synaptic connection loss", ``increase of hippocampal neurofibrillary tangles (NFTs) and plaques (NPs), which are strongly correlated with cognitive decline and are a hallmark feature of AD".

\item \textbf{Irrelevant claims}: Elicit's last statement (E4) is out of scope and discusses possible therapeutic strategy for AD. 

\item \textbf{Lack of depth}: Elicit's answer mostly attempts to describe links between source (``epigenetic dysregulation of neurotrophins") and target (``AD"). The mention of specific neurotrophic factors (BDNF) introduces a particular pathway but does not detail its mechanism. The reference to ``DNA hypermethylation and histone deacetylation" or to ``repressed chromatin state and reduced plasticity", as discussed above, are just very generic mechanisms that are not instantiated to answer the particular question at hand.
Cora's answer contains complex chains. For example, one of them links ``epigenetic dysregulation of neurotrophins" to ``increased promoter CpG methylation" to ``down-regulation of BDNF" to ``decrease synthesis of BDNF in neurons" to a ``reduction of synaptic proteins" to ``synaptic connection loss" that leads to ``neurodegeneration" and ultimately contributes to AD.

\end{itemize}

Cora provides an answer that is completely backed by evidence while also making cohesive connections across papers. It starts by clearly illustrating in what manner (CpG methylation) the down-regulation of BDNF, a neurotrophin, has been observed. Then, it illustrates that this dysregulation is associated with several diseases. It then goes on to illustrate the crucial function BDNF plays in several different cognitive functions and how a reduction in this neurotrophin leads to the disruption of several processes that are important in regular brain function. It also mentions important intermediate concepts such as NFTs and NPs which are unequivocally associated with AD pathology. Therefore, Cora provides a clear, coherent pathway that is backed by evidence while providing rich connections between concepts across different papers, successfully solving a multi-hop query.
\end{document}